\documentclass{article}

\usepackage{PRIMEarxiv}

\usepackage[utf8]{inputenc} % allow utf-8 input
\usepackage[T1]{fontenc}    % use 8-bit T1 fonts
\usepackage{hyperref}       % hyperlinks
\usepackage{url}            % simple URL typesetting
\usepackage{booktabs}       % professional-quality tables
\usepackage{amsfonts}       % blackboard math symbols
\usepackage{nicefrac}       % compact symbols for 1/2, etc.
\usepackage{microtype}      % microtypography
\usepackage{lipsum}
\usepackage{fancyhdr}       % header
\usepackage{graphicx}       % graphics
\graphicspath{{media/}}     % organize your images and other figures under media/ folder
\usepackage{algorithm}%
\usepackage{algorithmicx}%
\usepackage{algpseudocode}%
\usepackage{listings}%
\usepackage{multirow}%
\usepackage{xcolor}%

\title{Partitioned Neural Network Training via Synthetic Intermediate Labels
}

\author{
  Cevat V. Karadağ, Nezih Topaloğlu \\
  Department of Mechanical Engineering \\
  Yeditepe University \\
  Istanbul, Turkey\\
  \texttt{nezih.topaloglu@yeditepe.edu.tr} \\
}

\begin{document}
\maketitle

\begin{abstract}
The proliferation of extensive neural network architectures, particularly deep learning models, presents a challenge in terms of resource-intensive training. GPU memory constraints have become a notable bottleneck in training such sizable models. Existing strategies, including data parallelism, model parallelism, pipeline parallelism, and fully sharded data parallelism, offer partial solutions. Model parallelism, in particular, enables the distribution of the entire model across multiple GPUs, yet the ensuing data communication between these partitions slows down training. Additionally, the substantial memory overhead required to store auxiliary parameters on each GPU compounds computational demands.
Instead of using the entire model for training, this study advocates partitioning the model across GPUs and generating synthetic intermediate labels to train individual segments. These labels, produced through a random process, mitigate memory overhead and computational load. This approach results in a more efficient training process that minimizes data communication while maintaining model accuracy.
To validate this method, a 6-layer fully connected neural network is partitioned into two parts and its performance is assessed on the extended MNIST dataset. Experimental results indicate that the proposed approach achieves similar testing accuracies to conventional training methods, while significantly reducing memory and computational requirements. This work contributes to mitigating the resource-intensive nature of training large neural networks, paving the way for more efficient deep learning model development.
\end{abstract}

% keywords can be removed

\section{Introduction}\label{sec:intro}
In recent years, neural networks, particularly deep learning models, have emerged as powerful tools for solving complex tasks across various domains. These networks have grown exponentially in size and complexity, enabling breakthroughs in areas such as image recognition, natural language processing, and speech synthesis. However, this growth has also introduced a significant challenge for the average user: training very deep and large networks. For instance, Language Models like GPT-3, with billions of parameters, require substantial computational resources and specialized expertise to train effectively. Especially, the graphical processing unit (GPU) memory requirements of large networks pose a bottleneck in their training.

Recognizing the importance of scarce memory resources in deep learning, researchers and practitioners have been actively exploring various approaches to make the technology more accessible \cite{Sutton09,Shiram19,Fu2021,Yao2018,Meng2017TrainingDM}. A simple yet effective approach is to decrease the precision of the model parameters. Instead of using the 32-bit floating point data-type, BFLOAT16, which consumes 16 bits (8 bits for exponent and 7 bits for fraction) can be used \cite{URL_BFLOAT16}. Although this approach halves the memory consumption, it comes with decreased model parameter accuracy. In addition, the improvement it provides is limited: Using 8-bit data types, for example, would result in significant deterioration of accuracy. 

One promising avenue that decreases GPU memory consumption involves partitioning either the data, model or the pipeline, called parallelism \cite{jia2018data}. Data parallelism (DP) involves sharding the training batches into smaller subsets and distributing them across multiple users or devices, allowing collaborative training without the need for a centralized infrastructure. On the other hand, model parallelism (MP) focuses on vertically splitting the model into smaller components that can be trained independently and then combined to form a larger network. This approach enables distributed training, where different parts of the network can be processed simultaneously by different devices or individuals. Pipeline parallelism (PP), on the other hand, splits the network horizontally on different computing units. All parallelism methods have garnered significant attention and are considered hot topics in the pursuit of decreasing training computation and memory requirements of deep learning, as they offer potential solutions to mitigate the challenges posed by training very deep and large networks.

In data parallelism, the model can fit into a single GPU. The replicates of the model and training parameters are copied to multiple GPUs and the data is partitioned to the number of available GPUs. The forward and backpropagation is also performed parallel at each GPU separately. The gradients calculated at each GPU are then synchronized and the model parameters at each GPU are updated accordingly. While computationally effective, this method does not reduce memory requirement per device. Data parallelism is actively used in PyTorch and is known as distributed-data parallel (DDP) \cite{URL_DDP}. A thorough investigation of PyTorch's DDP can be found at \cite{Li20}. Although DDP is a versatile solution when the network fits into a single GPU, it is not alone applicable when the model size is greater than the GPU RAM. 

Pipeline parallelism can be advantageous when the model can be splitted horizontally, such as in image processing applications. A famous example is G-Pipe, proposed by Huang et al. \cite{huang2019gpipe}. They proposed a micro-batch pipeline parallelism strategy, by dividing the mini-batches into equal micro-batches. At the end of each mini-batch, gradients are accumulated and applied to update the model parameters. One major drawback of pipeline parallelism is the pipeline bubble, due to workload imbalance or inter-pipeline dependencies. In addition, batch-normalization across the pipelines requires special attention.

Model parallelism is a key strategy when the model and the training parameters, such as gradients, optimizer states and other temporary variables have a size greater than the GPU memory \cite{Castello19,Wanwu2022}. Rajbhandari et al. proposed a model and training parameter sharding method, motivated by achieving zero data overlap between GPUs \cite{rajbhandari2020zero}. The method is recently improved as Fully Sharded Data Parallel (FSDP), with a PyTorch implementation \cite{zhao2023pytorch}. Again the data is sharded among GPUs, but this time the model and the training parameters are sharded as well, with zero overlap. After performing forward and backward passes in a parallel manner, the gradients are synchronized centrally and the model update is then performed. Recently, Mlodozenie et al. proposed model and data parallelism, where each model partition is optimized into specific data shards \cite{mlodozeniec2023}. A loss function based on the data shards unseen by the subnetwork is defined. Akintoye et al. proposed partitioning the model layer-wise \cite{akintoye2022}, to minimize the communication overhead between devices and the memory cost during training. Layers are partitioned using the GPUs and then merged. Model partitioning is also common among graph neural networks \cite{liao2018graph} and within systems involving edge devices, such as Internet-of-Things networks (IoT) \cite{Na22,Oliveira19,parthasarathy2023}.

Other than data, model and pipeline parallelism, researchers proposed other methods for decreasing memory demand of training neural networks. Jain et el. proposed encoding the feature maps to save memory \cite{Jain18}. Wang et al. proposed superneurons that feature memory optimization by dynamically allocating memory for convolution workspaces \cite{Wang18}. The memory demand of adaptive optimization method is significant as well. Some studies decreased this memory footprint by simplifying the optimizer models \cite{anil19memoryefficient,shazeer2018adafactor}. These method have the risk of affecting model convergence.

In this study, a novel neural network training methodology, called Partitioned Neural Network (PNN) training, is developed. The network is partitioned into two or more subnetworks. Instead of training the entire network, these partitions are trained separately using synthetic data and original training data. Since each partition is trained separately, the communication of feedforward outputs and gradients to other partitions or a host device is eliminated. Therefore, the communication overhead is significantly reduced compared to other model parallelism methods. Additionally, separate training of partitions allows fine-tuning the training hyperparameters for each partition. This approach enables a reduction in total computational demand while preserving accuracy. Finally, training partitions of a model separately is also effective in alleviating a potential vanishing gradient problem, which is more likely to be observed in deep neural networks \cite{Kolbusz2017vanishing}.

\section{Proposed Method}

The method relies on the observation that the intermediate layer weights of a neural network inherently exhibit some degree of randomness. This randomness stems from the initial random initialization of weights within the neural network, combined with the iterative nature of the training process \cite{Maennel20}. Consequently, the ultimate value of weights in an intermediate layer will differ and be contingent upon the initial random weights \cite{franchi2021tradi}. Hence, we posit that upon partitioning a network, the individual segments can be independently trained using synthetic labels or feature maps generated through a random process.

\begin{figure}
\centering
\includegraphics[width=.7\linewidth]{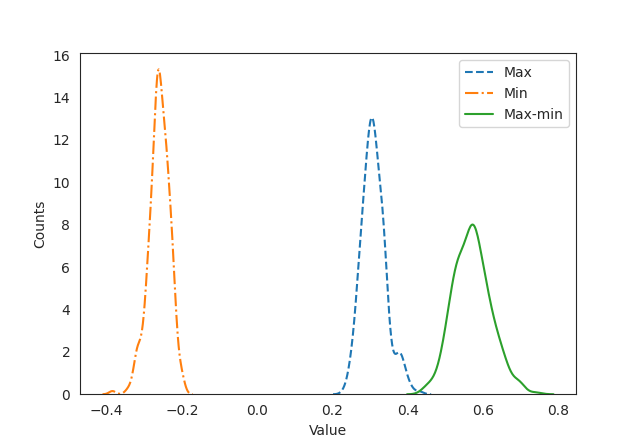}
\caption{The histogram of the maximum weight of the intermediate layer of a 3-layer network.}
\label{fig:histogram_of_maxes}
\end{figure}

To demonstrate the stochastic nature of network parameters after training, a simple fully-connected network is trained repetitively using the Modified National Institute of Standards and Technology (MNIST) dataset \cite{deng2012mnist}. The network has three layers, with the number of neurons at each layer as 100, 50 and 10. At each step, the weights are reinitialized randomly by recreating the network. The weight initialization is PyTorch's default initialization \cite{URL_torch_init_linear}. Training is performed in 15 epochs with a batch size of 256, and the model is saved after each of the 15-epoch training subroutine. After performing this full-training procedure 300 times, the histogram of three parameters are plotted: The maximum, minimum and the maximum minus minimum of the intermediate layer weights. The histograms are shown in Figure~\ref{fig:histogram_of_maxes}. The plots shows that even after training, the weights have some randomness, due to the random initial weights when the network is initiated. Therefore, the intermediate activation output of the network should also exhibit a significant amount of randomness, and it is expected that the network can be partitioned and trained using synthetic intermediate labels, which are generated using a random process.

\begin{figure}
\centering
\includegraphics[width=1\linewidth]{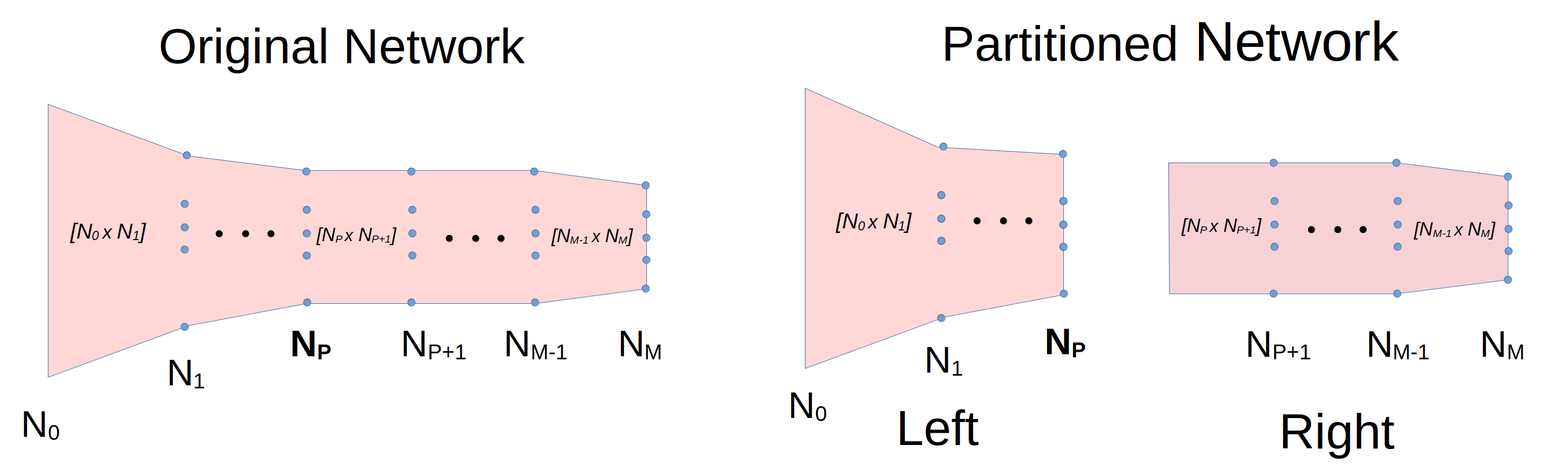}
\caption{The partitioning of an M-layer neural network.}
\label{fig:simplePNN}
\end{figure}

\begin{figure} [H]
\centering
\includegraphics[width=.7\linewidth]{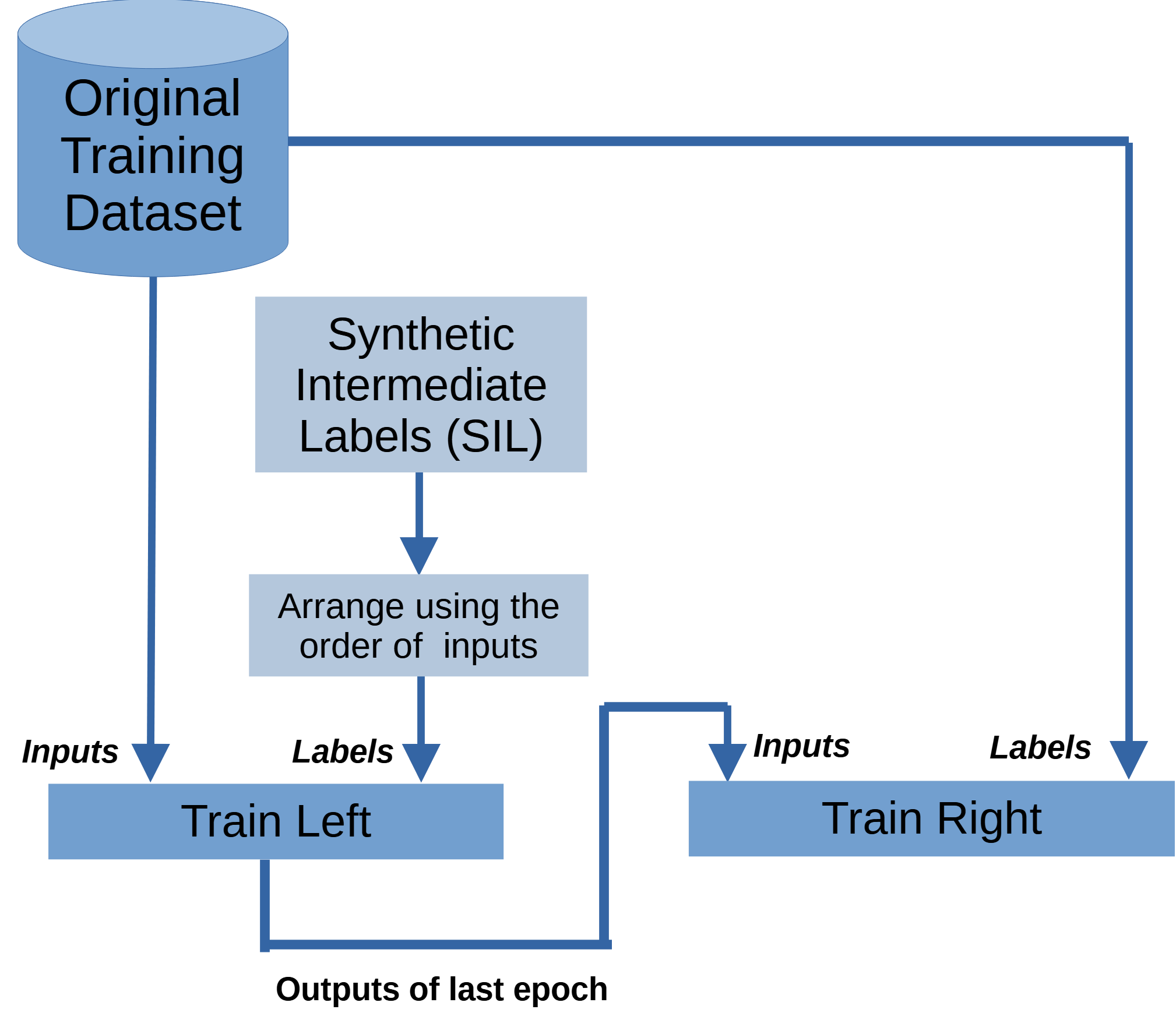}
\caption{The flowchart of the training algorithm for two partitions}
\label{fig:dataflowchart}
\end{figure}

The method starts by completely partitioning the model into two subnetworks, denoted as left and right partitions. This is conceptualized in Figure~\ref{fig:simplePNN}. Although partitioning into more than two subnetworks is possible, the method is explained assuming that the network is denoted into two subnetworks, for simplicity. The algorithm is shown in Figure~\ref{fig:dataflowchart}. The aim is to train the left partition first, without using the right partition. To achieve this, labels are synthetically generated for the left partition. These labels are called synthetic intermediate labels (SIL). Assuming that the number of neurons at the final layer of the left partition is $N_P$ and the number of classes is $M$, $M$ vectors of size $[N_P\times1]$ are created. The synthetic intermediate labels can be denoted by an N-by-M matrix ($SIL \in \mathbb{R}^{N_P\times M} $), where each column represents a class. The elements of the matrix are randomly generated using the $(0,1)$ uniform distribution and scaled using a parameter $\kappa$. In mathematical notation:
\begin{equation}\label{eq:SIL}
    SIL_{i,j} \sim \kappa \, U(0,1)
\end{equation}
where $U(0,1)$ is the $(0,1)$ uniform distribution and $i \in {1,2,...,N_P}$ and $j \in {1,2,...,M}$ are the row and column indices.

Using the original input of the training dataset and the SIL, ordered using the original training dataset, the left partition is trained by $N_L$ epochs without shuffling the input. In this stage, the right partition is not used at all. After $N_L$ epochs, training of the left partition is terminated and the final response of the left partition, i.e. the response at the last epoch, is stored.

In the second phase, the right partition is trained. The input data is the output of the final response of the left partition, stored previously. The labels of the original dataset are used as the labels of this training process. The training is performed by $N_R$ number of epochs. After the training of the right partition is finished, the training process is complete. The partitions can be joined after this stage, to use the network.

\begin{figure} [H]
\centering
\includegraphics[width=.85\linewidth]{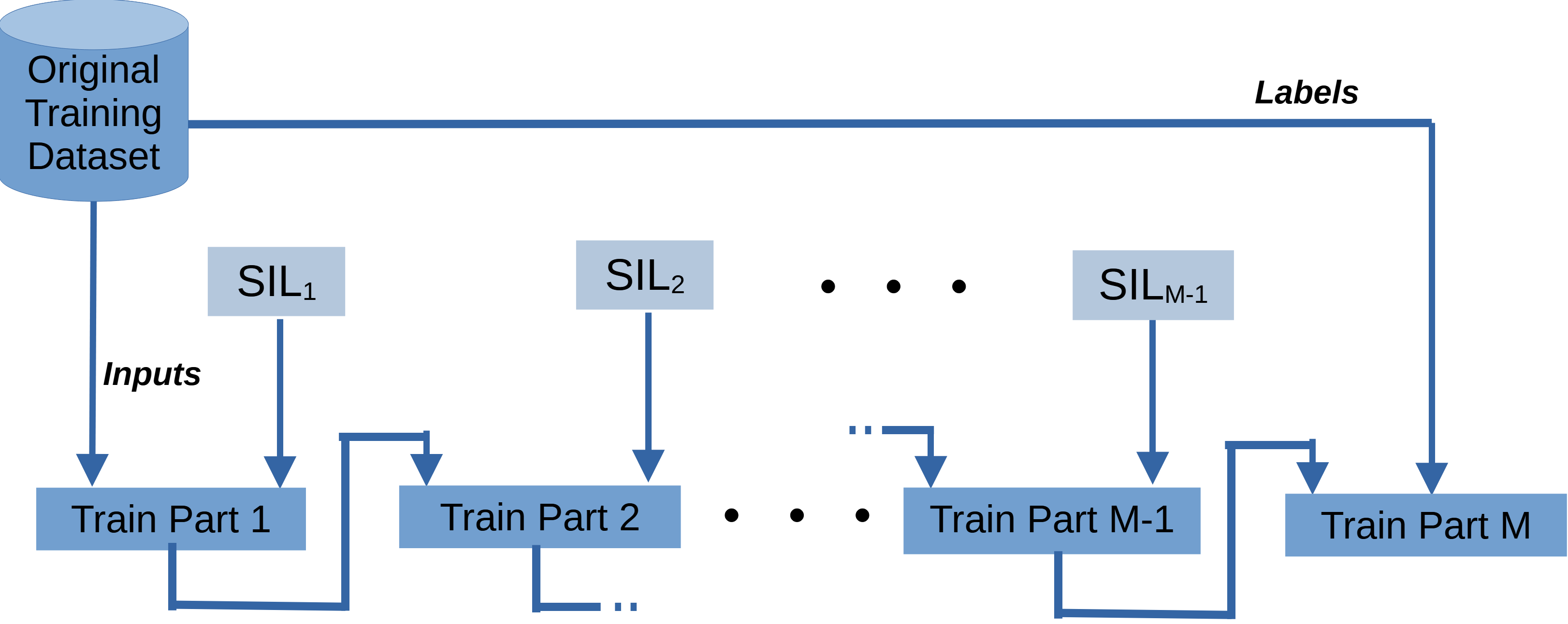}
\caption{The flowchart of the training algorithm for more than two model partitions.}
\label{fig:dataflowchart_all}
\end{figure}

The method can be readily expanded to cases where the model is partitioned into more than two subnetworks. This overarching algorithm is outlined in Figure~\ref{fig:dataflowchart_all}. In this scenario, a distinct Synthetic Intermediate Label (SIL) is necessary for each intermediate layer. The benefits of the proposed method remain applicable.

\begin{figure} [H]
\centering
\includegraphics[width=.85\linewidth]{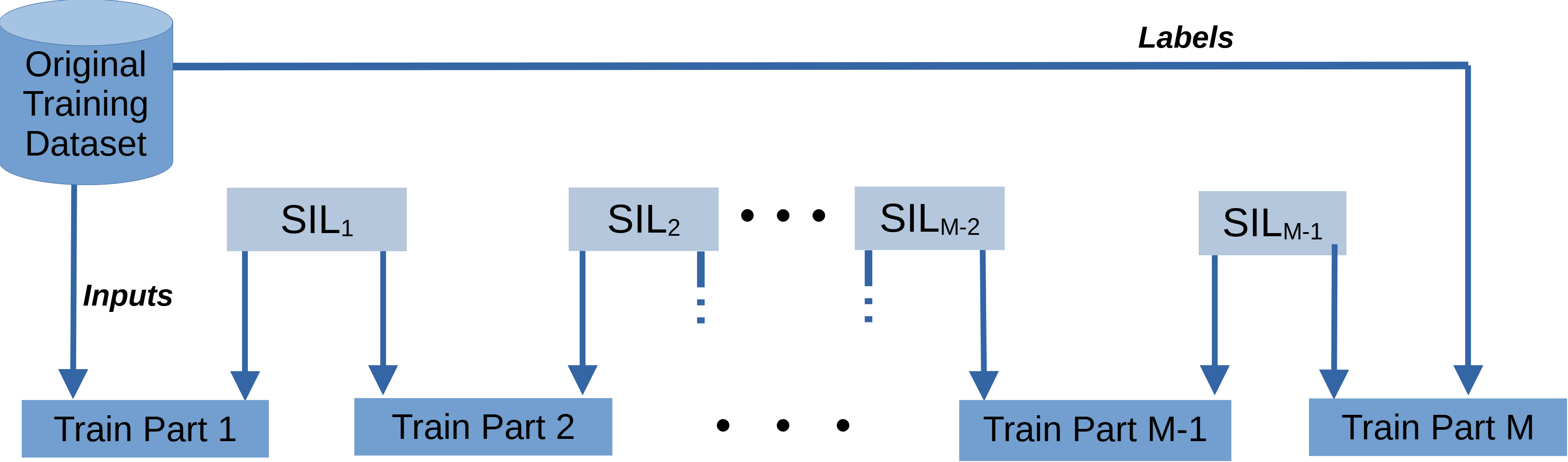}
\caption{The flowchart of the training algorithm with the intermediate layers trained using only SIL.}
\label{fig:dataflowchart_fully_separate}
\end{figure}

It is also feasible to break away from the sequential nature of training and simultaneously train each subnetwork, employing synthetic intermediate labels both as inputs and labels. The flowchart depicting this approach is presented in Figure~\ref{fig:dataflowchart_fully_separate}. This interesting architecture includes intermediate partitions, to be trained via inputs and labels generated by a random process. It was previously shown that it is possible to train a neural network with random data with zero training loss, ensuring that the number of epochs and the number of parameters are sufficient \cite{zhang2017understanding}. Nonetheless, this method mandates numerous epochs of training for each subnetwork to achieve an acceptable level of accuracy, thereby substantially increasing the computational load. As a result, this approach is considered impractical.

\subsection{Advantages to Standard Model Parallelism}

\underline{Communication overhead:} Conventional model parallelism entails extensive communication between devices. For instance, in a node with interconnected GPUs, both device-to-host and host-to-device communications are necessary to transmit responses, gradients, and updated model parameters \cite{Jain20,zhuang2022optimizing}. As the number of GPUs increases, the communication overhead intensifies, imposing an upper limit on overall performance \cite{rajbhandari2020zero}.

The proposed approach facilitates the training of each partition within its respective GPU. The sole communication overhead during training involves transferring the output of the preceding partition, which serves as the input for the current partition under training. This strategy mitigates communication overhead, presenting an efficient yet accurate training approach.

\underline{Device memory:} Both the proposed method and standard model parallelism share the advantage of distributing the model across multiple devices. Consequently, both approaches require memory to store model parameters and associated data (gradients, optimizer states, activations, etc.). Additionally, training data must be batch-transferred to device memory such as in GPUs. Similar case happens for L3 cache memory recently due to the availability of large L3 cache sizes on CPUs, such as AMD-EPYC-9684X with 1152 MB cache ~\cite{URL_epyc}. 
Diverging from model parallelism, the proposed training method is sequential. This enables the method to be applicable even when only a single device with memory smaller than the model size is available. In contrast, standard model parallelism would be significantly slow under such circumstances.

\underline{Computation demand:} Training partitions independently permits the customization of distinct training parameters for each partition, encompassing batch size, epochs, and learning rate. As an example, rather than subjecting the entire network to 40 epochs, one could allocate merely 5 epochs to the left partition while dedicating 80 epochs to the right partition. Such an approach can yield a remarkable reduction in total computation demand. Chapter~\ref{sec:results} demonstrates that the accuracy of the left partition converges within a few epochs, underscoring the advantage of this strategy.

%*****************************************************************************************************
%*****************************************************************************************************
\section{Implementation on Fully-Connected Networks}

The algorithm is applied to a fully-connected classification network. The extended MNIST (EMNIST) dataset \cite{cohen2017emnist}, balanced version, is used. The dataset includes 47 classes, including numbers, uppercase and lowercase letters. The inputs are $28\times28$ gray-scale images, flattened to $784\times 1$ vectors. 

The base (unpartitioned) network is a six-layer fully-connected network with bias. Following the input layer of size 784, the number of neurons at each layer are 80, 60, 60, 60 and 47. The network is partitioned at the third layer. Therefore the left partition has $80+60 = 140$ neurons whereas the right partition has $60+60+47 = 167$ neurons. However, since the input size is 784, the number of parameters of the left partition is considerably larger: $(784+1)\times80 + (80+1)\times60 = 67660$ parameters at the left partition and $(60+1)\times60 + (60+1)\times60 + (60+1)\times47 = 10187$ parameters at the left partition. The number of multiply-accumulate operations (MACs) are proportional to the number of parameters in a fully-connected network. Using a MACs counter library \cite{ptflops}, the MACs of the left and right partitions are calculated as $67800$ and $10307$, respectively. Therefore, training the left partition is more computationally intensive than training the right part. 

The synthetic intermediate labels are generated using Eq.~\ref{eq:SIL}, with $\kappa=10$. The size of the SIL is $60\times47$ where $47$ is the number of labels and $60$ is the number of neurons at the partition layer. There are $112800$ images in the training dataset. The label for each input image is a $60\times1$ vector, drawn from the $60\times47$ matrix of SIL. Since the inputs are not shuffled during training, the labels required to train the left partition ($112800$ labels of size $60\times1$) are arranged and loaded to the GPU as batches. Stochastic gradient method with learning rate of $0.01$ and momentum of $0.9$ is used for optimization. The activation at each function is the rectified linear unit (ReLU), except the final layer, where identity function is used. The batch size is set as 1410. The computations are computed on an AMD Radeon RX 7600 graphics card.

\section{Results and Discussion}\label{sec:results}

 In this section, the accuracy of the proposed method as a function of computation, in comparison with baseline training, is examined. For baseline training, the same  dataset is used to train the unpartitioned 6-layer network, employing identical optimization and batch size parameters.

The computation is calculated using the MACs of the left, right and unpartitioned models. The proposed method introduces communication overhead due to transferring the Synthetic Intermediate Label (SIL) to the GPU and conveying the outputs of the left layer as inputs to the right layer. Both of these transfers occur only once and are thus excluded from the analysis.

\begin{figure}[H]
\centering
\includegraphics[width=.8\linewidth]{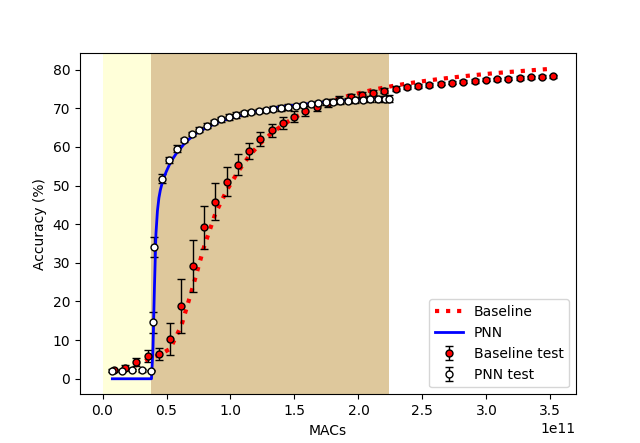}
\caption{The results for $N_L = 5$ and $\kappa=10$. The yellow and brown background shows the left training and right training, respectively.}
\label{fig:general_result}
\end{figure}

The outcomes for a specific scenario are depicted in Figure~\ref{fig:general_result}. The number of epochs for training the left, right and baseline models are $N_L = 5$, $N_R = 160$ and $N_B = 40$, respectively. The multiplier parameter $\kappa$ is set to 10. The acronym PNN represents the Partitioned Neural Network, signifying the proposed model. The complete training process is executed 10 times for both the baseline and PNN models, and error bar plots are employed. The circular markers denote the mean, while the error bars illustrate the $68\%$ range, equivalent to one standard deviation percentile. The yellow and brown backgrounds correspond to the training subprocesses of the left and right partitions, respectively.

The results unequivocally demonstrate that the proposed method achieves a testing accuracy comparable to conventional training (71.5\% for PNN and 76.2\% for the baseline) while utilizing significantly less computing power. Despite the right partition undergoing 160 epochs of training, its accuracy converges within a notably low number of epochs. Another pivotal finding is that the left partition, trained for just 5 epochs, contributes significantly to the reduction of the total computational load.

\begin{figure}[htb]
\centering
\includegraphics[width=.8\linewidth]{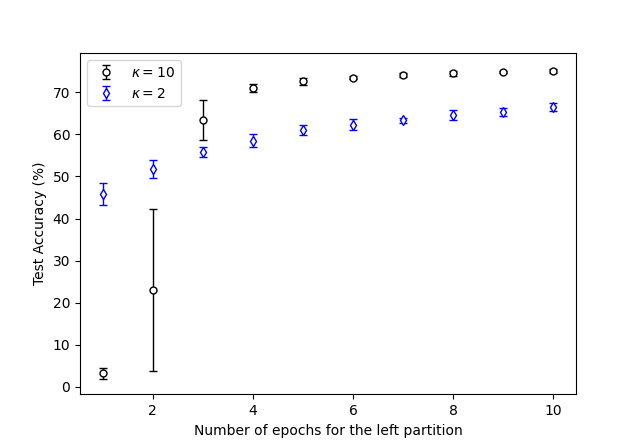}
\caption{The effect of the number of epochs in training the left partition, for $\kappa = 2$ and $\kappa = 10$. For all simulations $N_R = 160$.}
\label{fig:effect_of_NL}
\end{figure}

The outcomes suggest the possibility of fine-tuning parameters such as the number of epochs and the multiplier $\kappa$ to optimize efficient and accurate training. The effect of $N_L$ is given in Figure~\ref{fig:effect_of_NL}. The testing accuracy of the PNN is plotted as a function of $N_L$ for $\kappa = 2$ and $\kappa = 10$, while other parameters remain unchanged. The error bars represent one standard deviation percentile. The results underscore that training the left partition merely around 5 times is adequate to achieve convergent results. The effect of $N_L$ is more prominent when $\kappa$ is increased. Uneven training of the two partitions ($N_L = 5$, $N_L = 160$) represents a distinct advantage of PNN. By optimizing the number of epochs and other hyperparameters for each partition, the overall computational time can be greatly reduced. This is not possible in conventional model parallelism.

\begin{figure}[H]
\centering
\includegraphics[width=.8\linewidth]{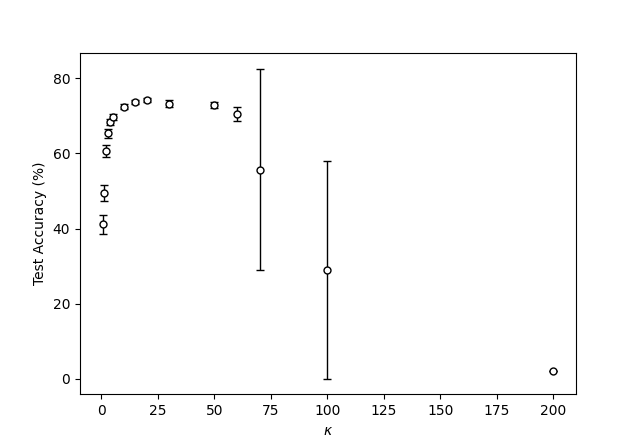}
\caption{The effect of the multiplier parameter, $\kappa$. The y-axis is the test accuracy after training the left partition by $N_L = 5$ epochs and the right by $N_R = 160$ epochs.}
\label{fig:effect_of_K}
\end{figure}

Figure~\ref{fig:effect_of_K} focuses on the impact of the $\kappa$ parameter on testing accuracy. It is evident that an optimal parameter value exists, leading to the highest testing accuracy. Notably, lower multiplier values correlate with diminished accuracy, while higher values exhibit a decrease in mean testing accuracy with higher variation, except very high values of $\kappa$, where training fails.

\begin{figure}[ht]
\centering
\includegraphics[width=.8\linewidth]{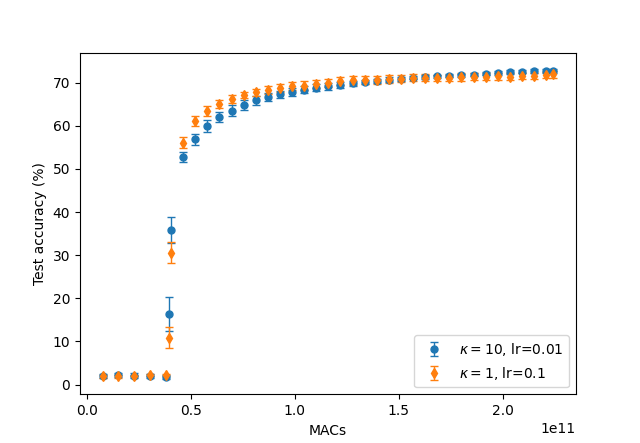}
\caption{The test accuracy of PNN with two different combinations of $\kappa$ and learning rate.}
\label{fig:effect_of_K_and_lr}
\end{figure}

Analogous to the learning rate, the multiplier parameter seems to exert a similar influence. Elevating $\kappa$ tends to amplify gradients, which are then multiplied with the learning rate in the weight update process. Therefore, it is hypothesized that the $\kappa$ and learning rate (lr) are analogous, as they seem to exert a similar influence. To substantiate this notion, test accuracy plots are generated for two scenarios: ($\kappa = 10$, $lr = 0.01$), and ($\kappa = 1$, $lr = 0.1$). As seen in Figure~\ref{fig:effect_of_K_and_lr}, the resulting accuracy vs. MACs plots for these cases display striking similarity with $R^2 > 0.99$, providing empirical confirmation of the previously mentioned hypothesis.

\section{Enhancing Accuracy through Recovery Epochs}
\begin{figure}[H]
\centering
\includegraphics[width=.8\linewidth]{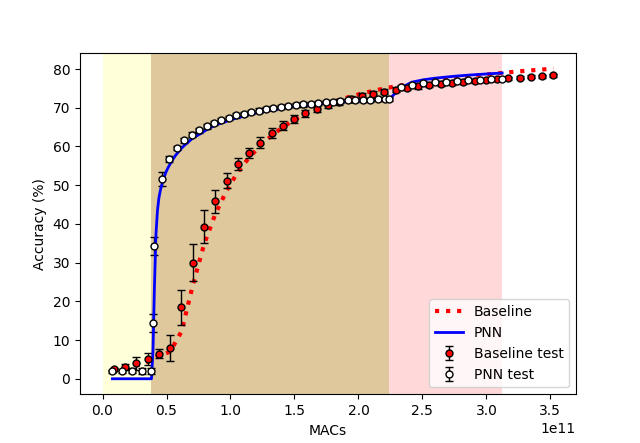}
\caption{The results after applying a recovery phase of 10 epochs. The yellow, brown and pink backgrounds depict the left training, right training and the recovery phase, respectively.}
\label{fig:general_result_with_recovery}
\end{figure}

Although the proposed method provides significant advantages, the results denote that the test accuracy is slightly lower than baseline network. This drawback can be partly alleviated by continuing training after finishing the proposed training method. The left part, which is trained only for a few epochs, are trained for a couple more epochs, while the right part weights are frozen. This approach, resembling a form of transfer learning, improves the accuracy. Figure~\ref{fig:general_result_with_recovery} shows the testing accuracy when a recovery phase of 10 epoch are applied, with the same configuration given in Figure~\ref{fig:general_result_with_recovery}. The mean testing accuracy which is $72\%$ after training the right partitions (depicted by the brown region), increased to $77.5\%$ following the recovery phase (depicted by the pink region). In this manner, the method not only offers initial efficiency benefits but also provides a pathway for accuracy refinement through extended training.

%*****************************************************************************************************
%*****************************************************************************************************
\section{Conclusions and Future Prospects}

In conclusion, this research introduces a groundbreaking approach to address the challenges posed by training resource-intensive neural networks. By synergizing model parallelism and synthetic intermediate labels, a novel methodology is devised that significantly enhances training efficiency without compromising model accuracy.

Empirical experiments on the EMNIST dataset validate the efficacy of the proposed approach. Partitioning a 6-layer neural network into segments, combined with synthetic intermediate labels, maintains testing accuracies comparable to conventional methods while reducing memory overhead and computational requirements. This advancement holds far-reaching implications, offering a practical solution to the resource constraints of modern deep learning. By optimizing training processes, this approach paves the way for more accessible development of advanced neural network models. 

In addition to its current contributions, the future landscape of research is poised to explore the broader application of this methodology beyond fully connected neural networks. This approach's adaptability holds promise for convolutional neural networks (CNNs), recurrent neural networks (RNNs), and transformer architectures. By tailoring the method to these diverse network types, researchers can further validate its generalizability and effectiveness across a wider range of deep learning models. This expansion into various architectures has the potential to revolutionize training efficiency across the entire spectrum of neural network paradigms, propelling the field toward more streamlined and accessible model development.\\

\section{Acknowledgements}
This preprint has not undergone peer review or any post-submission improvements or corrections. The Version of Record of this article is published in Multimedia Tools and Applications, and
is available online at \url{https://doi.org/10.1007/s11042-025-20666-9}”.

%Bibliography
\bibliographystyle{unsrt}  
\bibliography{manuscript}

\end{document}